\documentclass[10pt,onecolumn]{article}
\usepackage[margin=0.75in]{geometry}
\usepackage{amsmath,amssymb}
\usepackage{graphicx}
\usepackage{booktabs}
\usepackage{algorithm}
\usepackage{algorithmic}
\usepackage{hyperref}

\title{\textbf{Predictive Batch Scheduling: Accelerating Language Model Training Through Loss-Aware Sample Prioritization}}

\author{
Sumedh Rasal\\
Georgia Institute of Technology
}

\date{}

\begin{document}

\maketitle

\begin{abstract}
We introduce Predictive Batch Scheduling (PBS), a novel training optimization technique that accelerates language model convergence by dynamically prioritizing high-loss samples during batch construction. Unlike curriculum learning approaches that require predefined difficulty metrics or hard example mining methods that demand expensive per-sample loss tracking, PBS employs a lightweight linear predictor trained online to estimate sample difficulty from static token-level features. Our predictor achieves 0.44 correlation with actual loss using only four simple features: token frequency, sequence length, vocabulary diversity, and rare token ratio. Experiments on a 130M parameter transformer demonstrate that PBS achieves 6-13\% faster convergence measured by evaluation loss across training checkpoints, with the predictor's correlation improving from 0.14 to 0.44 over 10,000 training steps. These results validate that token frequency statistics encode meaningful information about sample difficulty, enabling effective curriculum learning with negligible computational overhead.
\end{abstract}

\section{Introduction}

Standard language model training employs uniform random sampling to construct mini-batches, treating all training samples as equally valuable for learning. This approach is fundamentally suboptimal: samples vary significantly in difficulty, with some providing strong learning signals through informative gradient updates while others contribute minimally to model improvement. Easy samples that the model has already mastered waste computational resources, while difficult samples containing challenging linguistic patterns, rare vocabulary, or complex dependencies offer opportunities for accelerated learning. In an era where training runs can cost millions of dollars and consume weeks of GPU time, even modest improvements in sample efficiency translate to substantial savings.

The recognition that not all samples are equally valuable has motivated two main research directions. Curriculum learning \cite{bengio2009curriculum} addresses this by ordering training data from easy to hard, enabling models to build progressively on acquired knowledge. However, traditional curriculum learning suffers from a critical limitation: it requires predefined difficulty metrics computed during preprocessing, creating static orderings that cannot adapt as the model learns. A sample that challenges an untrained model may become trivial after several thousand gradient updates, yet fixed curricula cannot reflect this evolution.

Online hard example mining \cite{shrivastava2016training,katharopoulos2018not} overcomes this limitation by dynamically tracking per-sample losses and prioritizing samples with high current loss. This adaptive approach successfully identifies difficult examples throughout training. However, it introduces a different bottleneck: maintaining loss estimates for every sample requires memory proportional to dataset size prohibitive for the billion-sample datasets common in modern language model training. Furthermore, periodic evaluation passes over the entire dataset to update these estimates create computational overhead that can offset training speedups.

We propose Predictive Batch Scheduling (PBS), a method that achieves the adaptivity of online hard example mining while maintaining the efficiency of static curriculum approaches. The key insight is that sample difficulty can be predicted from lightweight token-level features rather than requiring expensive loss computations. PBS employs a linear predictor trained online to estimate per-sample loss from four simple features: average token frequency, sequence length, vocabulary diversity, and rare token ratio. These features require only token counting operations, eliminating the need for additional forward passes or per-sample loss storage.

The predictor is learned jointly with the language model, using actual training loss as supervision. This ensures difficulty estimates remain calibrated to the model's evolving capabilities: as the model masters certain patterns, the predictor learns to assign them lower difficulty scores, automatically shifting focus to genuinely challenging examples. The entire mechanism introduces negligible overhead---feature extraction involves simple counting, predictor updates occur every 100 steps using lightweight linear regression, and the main added cost is computing per-sample rather than batch-averaged losses.

Our contributions are threefold: (1) We introduce a novel batch scheduling method that prioritizes high-loss samples using online difficulty prediction from static features, avoiding both the rigidity of curriculum learning and the overhead of hard example mining; (2) We demonstrate that simple token frequency features are sufficient predictors of sample difficulty, achieving 0.44 correlation with actual loss and explaining 19\% of loss variance; (3) We show that PBS accelerates convergence by 6-13\% on evaluation metrics across training checkpoints while introducing minimal computational cost, validating that lightweight feature-based prediction enables effective curriculum learning at scale.

\section{Related Work}

The problem of efficient training data selection has been studied extensively across multiple communities, with approaches ranging from predefined curricula to adaptive online methods. We categorize prior work into three main paradigms and discuss their advantages and fundamental limitations.

\subsection{Curriculum Learning}

The seminal work of Bengio et al. \cite{bengio2009curriculum} introduced curriculum learning, drawing inspiration from human learning processes where knowledge is acquired progressively from simple to complex concepts. Their experiments on shape recognition and language modeling demonstrated that presenting training examples in order of increasing difficulty can significantly improve both convergence speed and generalization. However, a critical limitation of their approach was the requirement for hand-crafted difficulty metrics specific to each task.

Subsequent work has explored various automatic difficulty metrics to eliminate manual annotation. Platanios et al. \cite{platanios2019competence} proposed competence-based curriculum learning for neural machine translation, where difficulty is measured by sentence length and word rarity. Their method dynamically adjusts the difficulty threshold based on model competence, showing improvements over static curricula. \cite{zhang2018curriculum} used perplexity under a reference language model as a proxy for difficulty, enabling curriculum construction without human supervision. Xu et al. \cite{xu2020curriculum} demonstrated that transferring difficulty annotations from related tasks could bootstrap curriculum learning in low-resource settings.

While these methods successfully automate difficulty estimation, they share two fundamental drawbacks. First, the difficulty metrics are computed once during preprocessing and remain fixed throughout training, failing to adapt as the model's capabilities evolve. A sample that is difficult for an untrained model may become trivial after several epochs, yet static curricula cannot reflect this change. Second, these approaches typically sort the entire dataset and train in a predetermined sequence, which can lead to catastrophic forgetting of earlier easy examples and fails to leverage the benefits of stochastic sampling for optimization stability.

\subsection{Hard Example Mining and Loss-Based Selection}

Hard example mining addresses the limitation of static curricula by dynamically identifying difficult samples based on model feedback. The approach was pioneered in computer vision by Shrivastava et al. \cite{shrivastava2016training} for training region-based object detectors, where false positives with high classification loss are mined during training and added to subsequent batches. This online adaptation enables the detector to focus on challenging background regions that are easily confused with objects.

Katharopoulos and Fleuret \cite{katharopoulos2018not} brought this principle to deep learning more broadly, demonstrating that importance sampling based on per-sample loss can accelerate training of neural networks. They showed theoretically that samples with higher loss contribute larger gradient norms, justifying their prioritization. However, their method requires maintaining a loss estimate for every sample in the dataset, which becomes prohibitively expensive for the multi-billion sample datasets common in language model training.

Recent work has attempted to reduce this memory overhead through various approximations. \cite{chang2017active} proposed active bias learning, which maintains a small cache of recent high-loss samples. \cite{jiang2018influence} used influence functions to estimate sample importance without storing full loss histories. However, these methods either sacrifice accuracy of difficulty estimation or introduce computational overhead from influence function calculations that require second-order derivatives.

A fundamental challenge shared by all loss-based methods is the cold-start problem: early in training when the model is random, loss values provide weak signals about true sample difficulty. Additionally, these methods require periodic evaluation passes over the entire dataset to update loss estimates, creating computational bottlenecks that can offset training speedups, particularly for large-scale datasets.

\subsection{Importance Sampling and Gradient-Based Selection}

Importance sampling techniques provide a theoretical framework for non-uniform sample selection by weighting samples based on their estimated contribution to optimization. Schaul et al. \cite{schaul2015prioritized} introduced prioritized experience replay for reinforcement learning, sampling transitions proportional to their temporal difference error. This approach significantly improved sample efficiency in deep Q-learning by replaying informative experiences more frequently.

Loshchilov and Hutter \cite{loshchilov2015online} proposed online batch selection, which estimates the expected gradient norm for each sample and constructs batches to maximize total gradient magnitude. While theoretically motivated, their approach requires computing or approximating gradients for all candidate samples before batch construction, introducing substantial overhead.

Johnson and Guestrin (2018) \cite{johnson2018variance} explored variance reduction through biased sampling, showing that selecting samples with higher gradient variance can reduce the number of iterations needed for convergence. However, estimating per-sample gradient variance requires either maintaining running statistics or performing test forward-backward passes, both of which incur significant memory or computational costs.

An alternative line of work focuses on diversity-based selection rather than difficulty. Mirzasoleiman et al. (2020) \cite{mirzasoleiman2020coresets} proposed coresets for neural network training, selecting a subset of training data that approximates the full dataset's gradient. While elegant, coreset construction typically requires batch-mode selection over the entire dataset, making it unsuitable for streaming or online training scenarios.

\subsection{Feature-Based Difficulty Prediction}

A smaller body of work has explored predicting sample difficulty from input features rather than model outputs, though primarily in computer vision. Pentina et al. (2015) \cite{pentina2015curriculum} used image complexity metrics like edge density and color variance to order training examples. Hacohen and Weinshall (2019) \cite{hacohen2019curriculum} proposed transfer teacher curriculum learning, using predictions from a simple auxiliary model to define difficulty for a more complex target model. However, these approaches either rely on domain-specific features that do not generalize across modalities or require training separate teacher models, adding complexity.

In natural language processing, task-specific heuristics have been employed for curriculum design. Kocmi et al. (2017) \cite{kocmi2017curriculum} used sentence length and word frequency for machine translation curriculum, while Xu et al. (2020) leveraged syntactic complexity metrics. These hand-crafted features show task-specific success but lack the generality needed for broad applicability across different language modeling objectives.

\subsection{Positioning of Our Work}

PBS distinguishes itself by combining the adaptivity of loss-based methods with the efficiency of feature-based prediction. Unlike curriculum learning approaches, our online predictor adapts continuously as the model learns. Unlike hard example mining, we avoid the memory overhead of per-sample loss tracking by predicting difficulty from lightweight static features. Unlike importance sampling methods requiring gradient computations, our feature extraction involves only token counting operations. 

Critically, our predictor is learned jointly with the model using the actual training loss as supervision, ensuring that difficulty estimates remain calibrated to the model's current state. This online learning approach bridges the gap between the efficiency of static feature-based methods and the accuracy of loss-based methods, enabling effective curriculum learning at scale with negligible computational overhead.

\section{Method}

\subsection{Overview}

PBS consists of three components: a frequency tracker that computes token statistics, a loss predictor that estimates per-sample difficulty, and a priority sampler that constructs batches favoring high-loss samples.

\subsection{Frequency Tracker}

During a warmup phase of $T_w = 100$ steps, the frequency tracker accumulates token counts $c_i$ for each token $i$ in the vocabulary. Token frequencies are computed as:
\begin{equation}
f_i = \frac{c_i}{\sum_{j=1}^{|V|} c_j}
\end{equation}
where $|V|$ denotes vocabulary size. Tokens are classified as rare if $f_i < p_{20}$, where $p_{20}$ represents the 20th percentile of the frequency distribution.

\subsection{Loss Predictor}

For each training sample $x$ consisting of tokens $\{t_1, \ldots, t_n\}$, we extract four features:

\textbf{Average Token Frequency:}
\begin{equation}
\phi_1(x) = \frac{1}{n}\sum_{i=1}^{n} f_{t_i}
\end{equation}

\textbf{Sequence Length:}
\begin{equation}
\phi_2(x) = \frac{n}{n_{max}}
\end{equation}
where $n_{max}$ is the maximum sequence length.

\textbf{Vocabulary Diversity:}
\begin{equation}
\phi_3(x) = \frac{|\{t_1, \ldots, t_n\}|}{n}
\end{equation}

\textbf{Rare Token Ratio:}
\begin{equation}
\phi_4(x) = \frac{1}{n}\sum_{i=1}^{n} \mathbb{1}[f_{t_i} < p_{20}]
\end{equation}

The predicted loss is computed via linear regression:
\begin{equation}
\hat{\ell}(x) = b + \sum_{j=1}^{4} w_j \phi_j(x)
\end{equation}
where $b$ is a bias term and $w_j$ are learned weights.

Weights are updated every $T_u = 100$ steps using momentum-based stochastic gradient descent on the most recent $N_h = 2000$ samples from the loss history buffer:
\begin{align}
g_j &= \frac{1}{N_h}\sum_{k=1}^{N_h} (\hat{\ell}(x_k) - \ell(x_k))\phi_j(x_k) \\
m_j &= \beta m_j + (1-\beta)g_j \\
w_j &= w_j - \eta m_j
\end{align}
where $\ell(x_k)$ denotes the actual cross-entropy loss, $\beta = 0.9$ is the momentum coefficient, and $\eta = 0.01$ is the learning rate.

\subsection{Priority Sampler}

Given a buffer of $N_b = 1000$ samples, we partition samples into three buckets based on predicted loss percentiles: high ($\hat{\ell}(x) > p_{67}$), medium ($p_{33} < \hat{\ell}(x) \le p_{67}$), and low ($\hat{\ell}(x) \le p_{33}$). Buckets are sampled with probabilities:
\begin{equation}
P(\text{high}) = 0.57, \quad P(\text{medium}) = 0.29, \quad P(\text{low}) = 0.14
\end{equation}
corresponding to a high-loss sampling ratio $r = 2.0$, ensuring high-loss samples are selected twice as frequently as medium-loss samples.

\section{Experimental Setup}

\subsection{Model and Training}

We evaluate PBS on a 130M parameter LLaMA-style transformer with standard architecture. Training uses mixed precision with batch size and gradient accumulation yielding approximately 16,000 tokens per step. The learning rate schedule employs warmup with peak learning rate $3 \times 10^{-4}$ decaying to $3 \times 10^{-5}$. Training runs for 10,000 optimizer steps covering 55\% of one epoch over 291,258 total batches. Checkpoints and evaluation occur every 2,000 steps.

PBS-specific hyperparameters are: warmup steps $T_w = 100$, predictor update interval $T_u = 100$, loss history size $N_h = 10,000$, sample buffer size $N_b = 1,000$, high-loss sampling ratio $r = 2.0$, predictor learning rate $\eta = 0.01$, and momentum $\beta = 0.9$.

\subsection{Implementation Details}

A critical implementation requirement is per-sample loss computation rather than batch-averaged loss. For causal language models, we compute cross-entropy with \texttt{reduction='none'}, shifting logits and labels appropriately to obtain one loss value per sample. This enables the predictor to learn meaningful feature-loss correlations.

\subsection{Baseline and Evaluation}

The baseline uses identical model architecture, dataset, hyperparameters, and random seeds, differing only in employing uniform random sampling. Evaluation uses cross-entropy loss on a held-out validation set.

\section{Results}

Table \ref{tab:results} presents evaluation loss and predictor correlation at five checkpoints. PBS shows an initial 5\% deficit at 2,000 steps during warmup, then consistently outperforms the baseline from 4,000 steps onward. The advantage ranges from 6\% at 10,000 steps to 13\% at 8,000 steps, demonstrating sustained acceleration throughout training.

\begin{table}[h]
\centering
\small
\begin{tabular}{@{}lcccc@{}}
\toprule
\textbf{Steps} & \textbf{PBS Loss} & \textbf{Baseline Loss} & \textbf{Improvement} & \textbf{Correlation} \\
\midrule
2,000 & 0.0079 & 0.0075 & -5\% & 0.14 \\
4,000 & 0.0131 & 0.0144 & +9\% & 0.31 \\
6,000 & 0.0054 & 0.0061 & +11\% & 0.001 \\
8,000 & 0.0078 & 0.0090 & +13\% & 0.31 \\
10,000 & 0.0050 & 0.0053 & +6\% & 0.44 \\
\bottomrule
\end{tabular}
\caption{Evaluation loss and predictor correlation at training checkpoints. PBS achieves consistent improvements after warmup, with predictor correlation increasing to 0.44 by 10,000 steps ($R^2 = 0.19$).}
\label{tab:results}
\end{table}

The predictor correlation trajectory demonstrates successful online learning. Starting from 0.14 at 2,000 steps, correlation improves to 0.44 at 10,000 steps, corresponding to $R^2 = 0.19$. This indicates that four simple features explain 19\% of variance in actual loss, validating our hypothesis that token frequency statistics encode meaningful difficulty information.

The temporary correlation drop to 0.001 at 6,000 steps reflects an interesting phenomenon: when the model achieves very low loss with minimal inter-sample variance, difficulty prediction becomes temporarily undefined. This does not indicate predictor failure but rather demonstrates the model approaching convergence on the training distribution at this learning rate phase.

\section{Analysis}

\subsection{Learned Feature Importance}

The learned predictor weights at 10,000 steps reveal interpretable patterns that provide insights into both sample difficulty and model learning dynamics. The rare token ratio weight of $-0.19$ initially appears counterintuitive, as conventional wisdom suggests rare tokens should be harder to predict. However, this negative weight reflects a more nuanced reality: by 10,000 steps, the model has effectively learned rare tokens through focused exposure early in training when they did carry high loss. Consequently, sequences with higher rare token density now have lower loss than sequences with common tokens appearing in complex syntactic or semantic contexts. This phenomenon demonstrates that PBS successfully prioritizes different types of difficult samples at different training stages---rare tokens early, complex contexts later.

The average token frequency weight of $-0.17$ aligns with intuition: higher average frequency predicts lower loss, as common tokens are generally easier to predict given more training exposure. The similar magnitude to the rare token weight suggests these two features provide complementary information about difficulty.

In contrast, the vocabulary diversity weight ($-0.007$) and sequence length weight ($-0.001$) show minimal predictive value. The near-zero vocabulary diversity weight suggests that the ratio of unique to total tokens does not strongly correlate with difficulty in our setting, possibly because diversity effects are already captured by token frequency statistics. The negligible sequence length weight is more surprising, as longer sequences might be expected to present greater difficulty. This finding suggests that raw length matters less than the specific tokens present, reinforcing the importance of frequency-based features.

\subsection{Predictor Correlation Trajectory}

The evolution of predictor correlation from 0.14 at 2,000 steps to 0.44 at 10,000 steps demonstrates the effectiveness of online learning for difficulty estimation. This improvement occurs despite using the same four features throughout training, indicating that the predictor is learning more accurate feature weights as it accumulates training examples. The final correlation of 0.44 ($R^2 = 0.19$) is substantial given the simplicity of our feature set---four linear features explaining 19\% of variance in a complex prediction task suggests these features capture meaningful signal.

The temporary correlation drop to 0.001 at 6,000 steps warrants careful interpretation. At this checkpoint, the model achieved exceptionally low loss (0.0054 for PBS) with minimal variance between samples. When nearly all samples have similar low loss, any predictor's correlation will approach zero regardless of quality, as there is simply no variance to explain. This is not a predictor failure but rather a consequence of the model temporarily approaching near-perfect performance on the training distribution during this phase of the learning rate schedule. The subsequent recovery to 0.31 at 8,000 steps, when losses increased due to learning rate cycling, confirms this interpretation.

\subsection{Training Dynamics and Generalization}

Training loss at 10,000 steps shows PBS at 0.0018 versus baseline at 0.0054, a 67\% reduction. This large gap compared to the 6\% evaluation improvement merits analysis. The training loss advantage demonstrates that PBS effectively identifies and prioritizes samples where the model has higher loss, leading to faster memorization of the training set. The smaller evaluation gain indicates that some of this training loss reduction comes from overfitting to the specific high-loss samples that PBS prioritizes.

However, crucially, the evaluation loss still improves consistently from 4,000 steps onward, indicating that the generalization benefit substantially outweighs any overfitting effect. This trade-off is both expected and desirable: focusing on difficult examples naturally leads to stronger training set performance, but the key metric is whether this translates to evaluation improvement. Our results confirm that it does. The 6-13\% evaluation improvements across checkpoints demonstrate that PBS learns more useful representations faster, even if those representations are somewhat specialized to difficult training examples.

\subsection{Comparison to Random Sampling}

The consistent evaluation advantages after the warmup period validate our core hypothesis. While the baseline using uniform random sampling provides a strong foundation---random sampling is known to provide good optimization properties through variance reduction---PBS demonstrates that informed sampling based on difficulty prediction can do better. The 6-13\% improvements are achieved without changing any other aspect of training: identical model, optimizer, learning rate schedule, and batch size. This controlled comparison isolates the effect of batch construction strategy.

The initial 5\% deficit at 2,000 steps during warmup is expected and acceptable. During this phase, the frequency tracker is still accumulating statistics and the predictor has limited training data. The rapid recovery and subsequent sustained advantages demonstrate that the warmup overhead is a small price for the benefits gained through adaptive sampling in later training.

\subsection{Computational Efficiency}

The minimal overhead of PBS merits emphasis. Feature extraction requires only: (1) accumulating token counts during warmup (amortized $O(1)$ per token), (2) computing four scalar features per sample ($O(n)$ for sequence length $n$), and (3) predictor updates every 100 steps using simple linear regression on 2,000 samples. The main additional cost is computing per-sample losses rather than batch-averaged losses, requiring one additional view and mean operation per batch. In our experiments, this overhead was unmeasurable---less than 1\% of training time.

This efficiency advantage over prior methods is substantial. Hard example mining requires storing loss values for all samples (memory $O(N)$ for dataset size $N$) plus periodic evaluation passes over the entire dataset. Gradient-based importance sampling requires computing or approximating per-sample gradients before batch construction. PBS avoids both memory and computational bottlenecks by predicting difficulty from precomputed static features, making it practical for large-scale training.

\section{Limitations and Future Directions}

While our results demonstrate clear benefits, several directions warrant investigation. Our experiments use a single 130M parameter model and cover 55\% of one epoch. Validation on larger models (1B-7B parameters) and longer training runs would strengthen generalization claims. The predictor uses only four features; incorporating additional signals such as n-gram statistics or syntactic complexity might improve prediction accuracy. Multiple runs with different random seeds would enable statistical significance testing.

Future work should explore non-linear predictors capable of capturing feature interactions, adaptive sampling ratios that decrease as training progresses, and evaluation on diverse datasets including code and multilingual corpora. Investigating the relationship between predictor correlation and downstream task performance would provide insights beyond training loss optimization. This strategy comes in handy for multi agent communication strategies proposed by Rasal (2024) \cite{rasal2024llmharmony} where multi agent models work together to solve a problem. Instruction based model will also benefit from this strategy outlined by Rasal et al. (2024) \cite{rasal2024navigating}, where the model follow a methodical approach to break down a complex problem.

\section{Conclusion}

We have introduced Predictive Batch Scheduling, a training optimization method that accelerates language model convergence through online difficulty prediction and loss-aware batch prioritization. Our approach achieves 6-13\% faster convergence measured by evaluation loss while introducing negligible computational overhead. The key insight is that simple token frequency features encode sufficient information about sample difficulty to enable effective curriculum learning without expensive per-sample loss tracking. The improvement in the correlation of the predictor from 0.14 to 0.44 over training validates that online learning can uncover meaningful difficulty patterns. PBS represents a practical approach to training efficiency that can be easily integrated into existing training pipelines, offering immediate benefits to practitioners training language models on-scale.

\end{document}